\documentclass[conference]{IEEEtran}

\ifCLASSINFOpdf

\else

\fi
\usepackage{algorithmic}

\usepackage{graphicx}

\usepackage[linesnumbered,ruled,vlined]{algorithm2e}
\SetKwInput{KwInput}{Input}             
\SetKwInput{KwOutput}{Output}
\usepackage{url}

\usepackage{booktabs}
\usepackage{amsmath}

\usepackage[latin9]{inputenc}
\usepackage{amsmath}

\usepackage{amssymb, array}

\usepackage[mathlines,switch]{lineno}
\usepackage{lipsum}

\makeatletter

 \usepackage[pscoord]{eso-pic}
\newcommand{\placetextbox}[3]{
 \setbox0=\hbox{#3}
 \AddToShipoutPictureFG*{ \put(\LenToUnit{#1\paperwidth},\LenToUnit{#2\paperheight}){\vtop{{\null}\makebox[0pt][c]{#3}}}
 }
 }
 \placetextbox{.23}{0.055}{\small{979-8-3503-3015-1/23/31.00~\copyright 2023 IEEE} }
\placetextbox{.23}{0.070}{\small{DOI:10.1109/ICCKE60553.2023.10326238}}

\let\old@ps@IEEEtitlepagestyle\ps@IEEEtitlepagestyle
\def\confheader#1{%
    \def\ps@IEEEtitlepagestyle{%
        \old@ps@IEEEtitlepagestyle%
        \def\@oddhead{\strut\hfill#1\hfill\strut}%
        \def\@evenhead{\strut\hfill#1\hfill\strut}%
    }%
    \ps@headings%
}
\makeatother

\confheader{%
        \parbox{20cm}{13th International Conference on Computer and Knowledge Engineering (ICCKE 2023), November 1-2, 2023, \\Ferdowsi University of Mashhad, Iran.}
}

\begin{document} 

\title{Developing Convolutional Neural Networks using a Novel Lamarckian Co-Evolutionary Algorithm}

\author{\IEEEauthorblockN{Zaniar Sharifi}
\IEEEauthorblockA{Faculty of Electrical and \\Computer Engineering\\
University of Zanjan\\
Zanjan, Iran 45371-38791\\
zaniar.sharifi@znu.ac.ir}
\and
\IEEEauthorblockN{Khabat Soltanian}
\IEEEauthorblockA{Faculty of Electrical and\\Computer Engineering\\
University of Zanjan\\
Zanjan, Iran 45371-38791\\
k.soltanian@znu.ac.ir}
\and
\IEEEauthorblockN{Ali Amiri\\}
\IEEEauthorblockA{Faculty of Electrical and\\Computer Engineering\\
University of Zanjan\\
Zanjan, Iran 45371-38791\\
a\_amiri@znu.ac.ir}
}

\maketitle

\begin{abstract}
Neural Architecture Search (NAS) methods autonomously discover high-accuracy neural network architectures, outperforming manually crafted ones. However, The NAS methods require high computational costs due to the high dimension search space and the need to train multiple candidate solutions. This paper introduces LCoDeepNEAT, an instantiation of Lamarckian genetic algorithms, which extends the foundational principles of the CoDeepNEAT framework. LCoDeepNEAT co-evolves CNN architectures and their respective final layer weights. The evaluation process of LCoDeepNEAT entails a single epoch of SGD, followed by the transference of the acquired final layer weights to the genetic representation of the network. In addition, it expedites the process of evolving by imposing restrictions on the architecture search space, specifically targeting architectures comprising just two fully connected layers for classification. Our method yields a notable improvement in the classification accuracy of candidate solutions throughout the evolutionary process, ranging from 2\% to 5.6\%. This outcome underscores the efficacy and effectiveness of integrating gradient information and evolving the last layer of candidate solutions within LCoDeepNEAT. LCoDeepNEAT is assessed across six standard image classification datasets and benchmarked against eight leading NAS methods. Results demonstrate LCoDeepNEAT's ability to swiftly discover competitive CNN architectures with fewer parameters, conserving computational resources, and achieving superior classification accuracy compared to other approaches.
\end{abstract}
\IEEEpeerreviewmaketitle

\section{Introduction}

The Convolutional Neural Networks (CNNs) have illustrated the dominance of various AI tasks, in the area such as object detection~\cite{1} and notably in image classification tasks on standard datasets such as AlexNet~\cite{2}, VGGNet~\cite{3} to very deep CNNs ResNet~\cite{4} which broken the state-of-the-art classification accuracy records in recent years. These successful dominance motivated researchers to propose complex and Deep Neural Networks (DNNs). However, these architectures are all designed manually by experts through a trial-and-error process precipitated handcrafting CNNs has become much more complicated on enormous datasets due to high dimensional architecture search space. Apart from difficulty, even the experts require considerable resources and time to create well performing architecture. 

In order to decline demanding development costs, a novel idea of Neural Architecture Searching (NAS) has emerged. In this specific domain, the disciplines of Reinforcement Learning (RL) and Evolutionary Computation (EC) are indeed well-established machine learning methodologies that have gained popularity and widespread adoption. The works of researchers~\cite{5,6} have provided compelling evidence demonstrating the effectiveness of RL-based approaches in producing CNN architectures that can outperform handcrafted counterparts in terms of performance.

Despite the promising outcomes achieved by the aforementioned RL-based algorithms on various datasets, their utilization demands substantial time and computational resources. Notably, in the study by~\cite{5}, it took 28 days and the utilization of 800 K40 Graphical Processing Units (GPUs) to explore and identify the optimal architecture. Such resource requirements are unfeasible for individual researchers and even certain organizations. In contrast, employing EC methods for the automated evolution of CNNs has demonstrated promising performance with a comparatively reduced reliance on GPUs compared to RL-based algorithms, e.g.~\cite{7,8}. One of the ways to reduce computation cost for both EC and RL methods is to prevent thorough training of each architecture during the architecture search by reducing the number of epochs for training every candidate solution.

Both EC and RL methods achieve high-performing CNNs by evaluating a large number of candidate solutions, which incurs significant computational costs and training time. This is due to the expansive nature of the architecture search space, although it can be constrained to mitigate computational expenses. EC methods have the capability to incorporate domain-specific constraints and prior knowledge, such as architectural principles, enabling further refinement of the search space. Consequently, EC methods offer a more efficient approach for discovering effective CNN architectures compared to RL methods. In this paper, we restrict the architecture search space by
focusing on CNN architectures that include only two fully connected layers as a classification segment.

In addition to the computational power challenges encountered by NAS methods, a significant hurdle lies in the simultaneous evolution of millions of weights and deep architectures. This process entails two intricate issues. Firstly, perturbing only a few weights per generation would necessitate numerous generations to fine-tune all weights. Conversely, perturbing a substantial number of weights in every generation could lead to excessively drastic modifications in the functionality of the DNN, impeding systematic search progress. This issue implies that the functionality of DNN remains unaltered, as though all the weights are initialized randomly~\cite{9}. Secondly, due to the pronounced sensitivity disparities among the DNN's parameters, altering a weight in the layer immediately following the inputs produces cascading effects that propagate through subsequent layers, unlike perturbations in layers closer to the output. Therefore, most of the recent NAS methods solely concentrate on discovering neural architectures and employ the Stochastic Gradient Descent (SGD) algorithm for optimizing hyperparameters~\cite{5,6,7,8,10,11,12,13}. Notable, the last layer is an excellent candidate for evolution due to its unique characteristics. The last layer does not propagate sensitivity during the feedforward process, and it plays a crucial role in transferring the true loss function to previous layers through backpropagation, provided that its weights are appropriately tuned.

In this paper, a graph-based Genetic Algorithm (GA) method, entitled LCoDeepNEAT, will be proposed to co-evolve CNN architectures and their last layer weights. LCoDeepNEAT is based on the CoDeepNEAT method~\cite{10} and aims to discover CNN architectures with minimum complexity. In addition, it employs the Lamarckian idea to neglect randomly initializing the network weights before training. Instead, it inherits the already adjusted weights of an ancestor network during the evolutionary process. The proposed method indicates how GA operators with the Lamarckian idea work and speed up evolutionary algorithms.
Our contributions can be summarized as follows:
\begin{itemize}
\item Defining a new genotype representation for evolving CNN architectures and tuning of their last layer weights from classification segment.

\item Calculating more accurate fitness that is tuned weights of the last layer are preserved by Lamarckian idea in genotype and are reused for next evaluations.

\item Speeding up the evolutionary process by constraining the search space to discover architectures with two fully connected layers as a classifier.

\item Generating solutions which outperform all hand-crafted CNNs and some NAS methods.
\end{itemize}

\section{Related Works}
Several search strategies are employed to traverse the CNN architecture space, encompassing Bayesian Optimization (BO), Reinforcement Learning (RL), and evolutionary methodologies.

BO is one of the successful algorithms in NAS. It is the first algorithm for tuning neural networks to win on competition datasets against state-of-the-art human expert neural network designing~\cite{11}. Recently, in~\cite{12}, it applies the BO algorithm for hyper-parameter architecture optimization.

RL methods have been widely investigated for the autonomous generation of CNNs. The agent, typically implemented as a recurrent neural network (RNN), generates architectures through its actions, with each action resulting in the discovery of a novel architecture within the search space. The accuracy of the architecture can be considered as the reward of the algorithm to update the agent. Despite achieving promising results on datasets such as CIFAR-10 and Penn Treebank (PTB)~\cite{5}, RL-based algorithms require significantly more time and computational resources compared to evolutionary algorithms~\cite{13}. For instance, the NAS method entailed the utilization of 800 GPUs over 28 days to identify an efficient CNN architecture for CIFAR-10, whereas the Genetic CNN achieved similar performance using 17 GPUs in one day. Therefore, Evolutionary Algorithms (EAs) are preferred due to limited availability of computational resources for researchers.

Over an extended period, EC algorithms have been employed by numerous researchers for the evolution of neural architectures, including the optimization of neural network weights. EAs originally being used solely to evolve the weights of a fixed architecture, since then in~\cite{14} proposed advantages of evolving both neural architecture and its weights that called Neuro-Evolution of Augmenting Topologies (NEAT). The NEAT method proposed three different mutations: (i) Add a connection between existing nodes, every connection plays the role of weight (ii) insert a node between two nodes while splitting an existing connection (iii) modify weights. Besides, it makes a breakthrough by providing a historical marketing mechanism; for preventing the problem of the same individual in terms of architecture, in the other word, it causes adding a new structure without losing track of which gene is all over the evolutionary process, and strategy of speciating; for promoting diversity that is known as fitness sharing mechanism. Owing to empirical promising results on different real problems, the NEAT method has gained attention from many researchers and various approaches have been developed based on the principle of the NEAT method such as HyperNEAT~\cite{15}, ES-HyperNEAT~\cite{16} and in the context of DNNs CoDeepNEAT method was developed. The CoDeepNEAT method was inspired by Hierarchical SANE~\cite{17}, ESP~\cite{18}, CoSyNE~\cite{19} and also the successfulness of well-known CNN architectures approaches such as~\cite{20}, using repetitive modules. CoDeepNEAT is a graph-based method that co-evolves two populations, called module and blueprint population. The blueprint individual is a Directed Acyclic Graph (DAG) whose every blueprint gene is a module individual. In addition, each module individual is a DAG, which represents a small CNN architecture. During fitness evaluation, all small DNN architectures of blueprint individual's gene assemble to create a larger CNN architecture.

Particle Swarm Optimization (PSO) is another EC algorithm that achieves potential for optimizing neural network architectures. Recent advancements in PSO algorithms, such as IPPSO~\cite{21}, and PSOCNN~\cite{22} have shown promising results. However, empirical evidence suggests that the architectures discovered by these algorithms remain complex and comprise millions of parameters. Consequently, there is a need to augment the efficacy of the architecture search of PSO algorithms.

In the context of DNN, NAS methods solely concentrate on neural architecture searching and neglect evolving their weights, owing to the high dimension of architecture search space, containing millions of weights and hyperparameters. Therefore, they often use back-propagation for optimizing the architecture's weights. In this paper, a graph-based genetic algorithm method is depicted to evolve CNN architectures and their last layer weights. LCoDeePNEAT employes the SGD algorithm to evaluate a candidate solution during evolution and transfers the acquired weights from the final layer to the corresponding section within the network's genotype. This strategy triggers an improvement in the classification accuracy of Individuals throughout the evolution and yields optimal architectures.

Differences with the CoDeepNEAT:
\begin{enumerate}
\item Having new representation at the Individual level; we propose a direct representation to encode CNN architecture and their last layer weights, while CoDeepNEAT proposed a direct representation just for encoding CNN architectures (Section 3).

\item  Partially evolving weights of the last layer of Individuals; at the Individual level, every CNN architecture and their last layer weights are co-evolved (Section 3.4).

\item Restricting architecture search space; LCoDeepNEAT meticulously discovers CNN architectures with two fully connected layers as the classifier segment (Section 3.1).

\item  Using the Lamarckian idea to tune the last layer weights of Individuals; the Lamarckian idea provides the possibility to inherit the already adjusted weights of an ancestor network during the evolutionary process. Our method returns the trained classifier layers of a CNN architecture into its Individual genome. The reproduction step transfers this set of weights (with possible modification) to the resulting offspring (Section 3.3).

\item Every genome is initialized with at least two genes; In CoDeepNEAT every genome is initialized with a gene, but in LCoDeepNEAT every genome is initialized with a number between two and four genes.
\end{enumerate}
\section{Proposed Method}
In this part, we first introduce representation and then the procedure for mapping the genotype to its corresponding phenotype, which manifests as a real CNN architecture in the context of LCoDeepNEAT. Secondly, the architecture search space is explained. Finally, the framework of LCoDeepNEAT is expounded in Subsection 3.2, followed by a comprehensive delineation of the essential algorithmic steps, spanning from Subsection 3.3 to Subsection 3.4.

 This paper presents a graph-based method comprised of two populations, referred to as 'module' and 'Individual,' aimed at discovering optimal CNN architectures. Our proposed method draws inspiration from the CoDeepNEAT method in terms of the structure of populations and their associated operators. Figure 1.a and Figure 1.b depict the genotypes of the Individual population and the module population, respectively. The Individual genotype consists of a DAG and a variable-length list (refer to Figure 1.a). In this graph, each node (or gene) is replaced by a module sourced from a subpopulation known as 'species,' with each module representing a small CNN. Moreover, every Individual has two constantly nodes as the classification segment. It is worth noting that orange nodes in the graph indicate fully connected layers. The variable-length list represents the weights of the last layer of the CNN architecture. Similarly, the module genotype is also a DAG where each node represents a layer in the CNN architecture. Broaldy speaking, the genotype of each module comprises a collection of filtering layers within a CNN architecture.

For the fitness evaluation of every candidate solution, all the genes of the Individual are replaced with the corresponding module to create a large CNN architecture (see Fig.2). Subsequently, two fully connected layers are appended to the terminal position of each Individual as a dedicated classification segment. To elaborate further, it is noteworthy that every Individual within the population consistently incorporates these two fully connected layers at its conclusion. Following this augmentation, the weights of the final layer are initialized in accordance with the variable-length list.

\begin{figure}[ht]
	\centering
	\includegraphics[width=\linewidth]{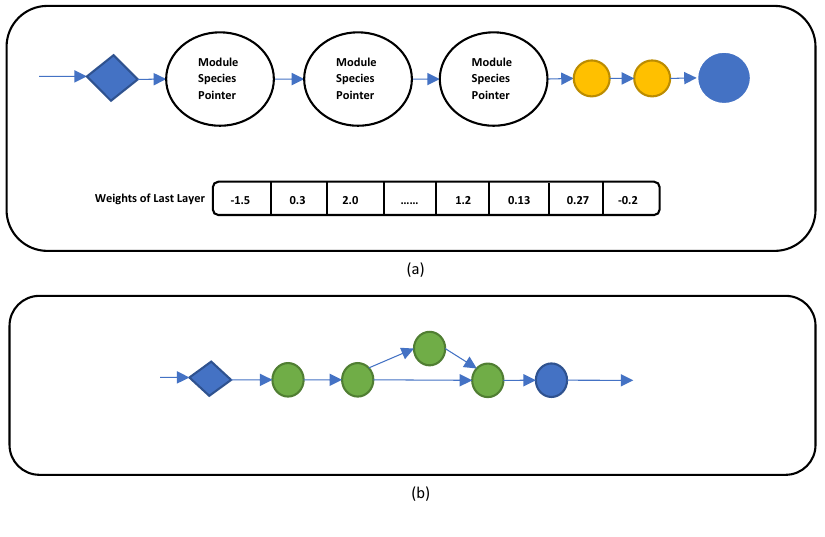}
	\caption{Fig 1.a shows Individual representation. Fig 1.b shows Module 
		representation. The blue nodes indicate input and output of the representation.}
        \label{fig-h}

\end{figure}
\begin{figure}[ht]
	\centering
	\includegraphics[width=\linewidth]{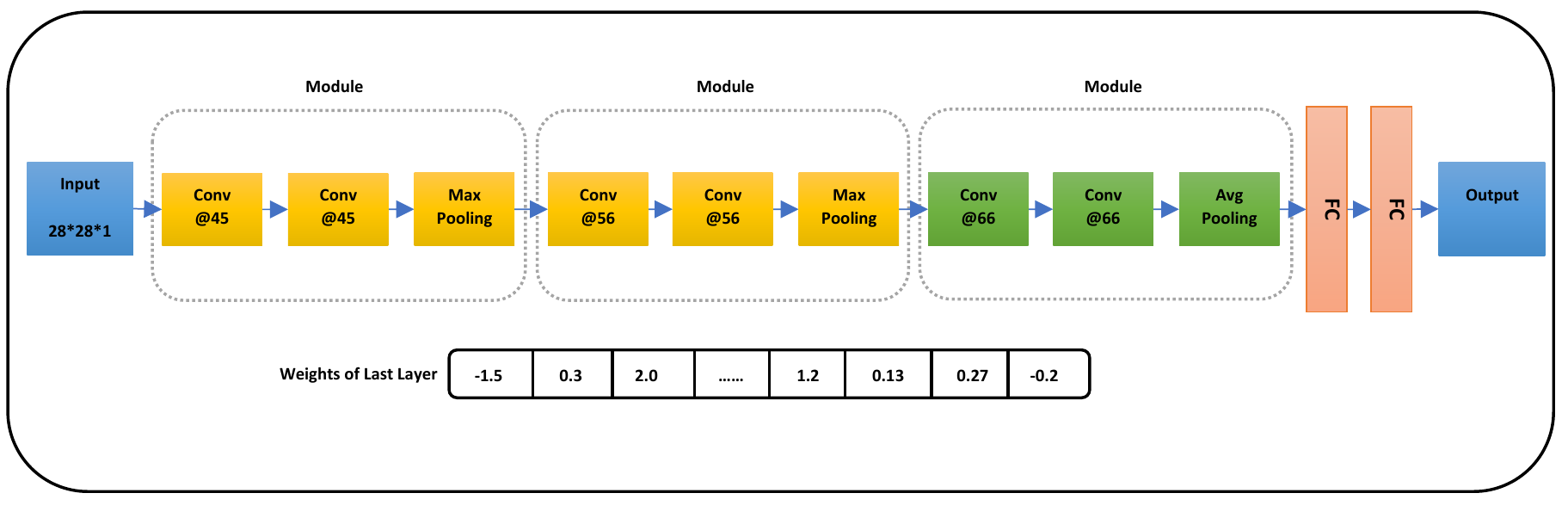}
	\caption{An example of the proposed method for representing a CNN.}
         \label{fig-h1}
\end{figure}
\subsection{Architecture Search Space}

The main aim of NAS is to find the best neural architecture in neural architecture search space, ${S_A}$, that encompasses all possible architectures on a given problem. Every point in this space denotes, ${a}$, a neural network architecture. By defining a performance criterion for neural network architecture (accuracy and network complexity), the performance of all possible networks forms a discrete level in this surface. Optimal architecture design is equivalent to finding the maximum point at this surface. There are several characteristics of such a surface, which illustrate that the process of NAS is highly complicated and time-consuming.

\begin{itemize}

\item The surface is infinitely enormous; because the number of nodes, connections and layers have high boundaries.

\item The surface is non-derivative; Since the change in the number of nodes and connections is discrete.

\item The surface is complex and noisy; the mapping between network architecture and its efficiency is done indirectly.

\item The surface is deceptive; the same architecture may be attributed to a different fitness.
\item The surface is multi-dimensional; different architectures may have the same functionality.

\end{itemize}
CoDeepNEAT and LCoDeepNEAT concentrate on finding CNN architectures, ${S^{\prime}}_{A}$, included sub neural networks that could be defined repetitive module. In addition, LCoDeepNEAT more constrained architecture search space, denoted as \text{S}$_{LCoDeepNEAT}$, which is a subset of ${S^{\prime}}_{A}$. Specifically, it targets architectures that comprise just two fully connected layers for classification. This can be expressed as:
\begin{equation*}
{S_{LCoDeepNEAT} \subseteq S_A^{\prime}}
\end{equation*}
CoDeepNEAT, on the other hand, explores the entire architecture search space without imposing any constraints. It operates within the full search space ${S^{\prime}}_{A}$. This is represented as:
\begin{equation*}
{S_A^{\prime}{=}S_{A_{CoDeepNEAT}}}
\end{equation*}
\subsection{Algorithm Overview}
\begin{algorithm}
	\KwInput{A set of minimum neural architecture, population size, maximum generation count, and the image dataset designated for classification purposes.}
	\KwOutput{The discovered the best CNN architecture in terms of accuracy.}
	\caption{framework of LCoDeepNEAT}\label{alg:one}
	\While{ $t$ $<$ the maximal generation number }{
		Evaluate the fitness of each assembled individual in $P_{{ind}_t}$ 
		and update 
		fitness of module in $P_{m_t}$ related to every used individual\;
		$M$ $\gets$ Select parent solutions from species in 
		$P_{{ind}_t}$ with tournament selection\;
		$F_t$ $\gets$ Generate offspring from $M$ with proposed mutation and 
		the 
		crossover operators\;
		$P_{m_{t+1}}$ $\gets$ survival selection on $P_{m_t}$ $\cup$ 
		$F_t$\;
		$P_{m_{t+1}}$ $\gets$ Speciation on $P_{m_{t+1}}$\;
		$S$ $\gets$ Select parent solutions from $P_{{ind}_t}$ with 
		tournament selection\;
		$Q_t$ $\gets$ Generate offspring from $S$ using the 
		proposed mutation and the crossover operators\;
		$P_{{ind}_{t+1}}$ $\gets$ survival selection on 
		$P_{{ind}_{t}}$ 
		$\cup$ 
		$Q_t$\;
		$P_{{ind}_{t+1}}$ $\gets$ Speciation on 		
		   $P_{{ind}_{t+1}}$\;
		${t}$ $\gets$ ${t+1}$\; 
    } 
\textbf{Return} \text{the individual having the best fitness in} $P_{{ind}_t}$\;
\end{algorithm}
Algorithm 1 illustrates the structural framework of the proposed algorithm. Firstly, two populations, one individual and one module, are initialized based on the proposed variable-length gene encoding strategy (line 1). Then, a generation counter is initialized to zero (line 2). After that, the process of evolution begins by assembling modules from individuals to map to CNN architectures where the weights are randomly initialized. The evaluation of the CNN architecture is conducted on the designated dataset to yield effective outcomes, continuing until a predetermined termination criterion, reaching the maximum number of generations, is satisfied. (lines 3-14).  Finally, the algorithm identifies the optimal CNN architecture (line 15) for the classification task of the given image dataset. All candidate solutions are assessed based on a fitness measurement during the evolutionary process. The fitness measurements of the individuals are then assigned to the modules that are the genes of the individuals (line 4). Then, the best modules of the module population are selected as a parent based on the tournament algorithm for generating new offspring (line 5). Next, the selected parents generate new offspring based on the proposed crossover and mutation strategy (line 6). Later, the generated offspring, the representation of every species, and the best module surviving from the current module population are selected for the next generation (line 7). After that, the new population of modules is speciated (line 8). Subsequently, the best individuals based on the tournament algorithm are selected from species as parents for reproduction (line 9). Following this, new offspring of individuals are generated with the designed mutation and crossover genetic operators (line 10). Then, the generated offspring, the representation of every species, and the best individual surviving from the current individual population are selected for the next generation (line 11). After that, the new offspring and survival individuals are speciated for the next generation, this action causes semi-structures to be collected in a group called species (line 12). Then, the generation counter, $t$, is incremented by one and the evolution continues until the counter exceeds the predefined maximal generation. Finally, the best CNN architecture is discovered for classifying the given image dataset. In the following subsections, the key steps of Algorithm 1 are elaborated in detail.

\subsection{Fitness Evaluation}
\begin{algorithm}
	\KwInput{The population of individual \text{P}$_{ind}$, the population of module \text{P}$_{m}$, Number of 
	neural networks selected from population of individual \text{num}$\_$network, the training 
	epoch number $k$ for measuring the accuracy tendency, the training set \text{D}$_{train}$ and the 
	fitness evaluation dataset \text{D}$_{val}$.}
	\KwOutput{Neural Networks selected from population of individual with fitness 
	$P_{t}$.}
	\caption{Fitness Evaluation }\label{alg:two}
	i=0, best$\_$fit = 0, avg$\_$fit= 0, best$\_$CNN$\_$architecture = Noun\;
	\While{i $<$ num$\_$network }
	{
		ind $\gets$ randomly selected an individual from \text{P}$_{ind}$\;
		ind.age + = 1\;
		$C$ $\gets$ mapping the ind to a CNN architecture\;
		$C$ $\gets$ initialized last layer of $C$ from the list of weights and bias of 
		the ind\;
		\ForEach{epoch in the given training epochs determined ${k}$}
		{
			fitness, architecture $\gets$ Train the $C$ on \text{D}$_{train}$ and calculate the 
			classification accuracy on \text{D}$_{val}$\;
		}
		\textbf{end}\;
	assign trained-weights and trained-bias of the last layer of the $C$ to list of weights 
	and bias of the ind\;
	\If{fitness $>$ best$\_$fit }
	{
		best$\_$fit $\gets$ fitness\;
		best$\_$architecture $\gets$ architecture\;
	}
\textbf{end}
	avg$\_$fitness +=fitness\;
	\ForEach{module in ind }
	{
		module.fitness +=fitness\;
	}
\textbf{end}
	$i \gets i +1$\;
avg$\_$fitness $\gets$ num$\_$network\;
\ForEach{module in \text{P}$_{m}$ }
{
	\If{module.age==0 }
	{module.fitness $\gets$ avg$\_$fitness+0.01\;}
	\textbf{end}
}
\textbf{end}
\ForEach{individual in \text{P}$_{int}$  }
{
	\If{individual.age==0  }
	{individual.fitness $\gets$ avg$\_$fitness+0.01\;}
	\textbf{end}
}
\textbf{end}
	}
\textbf{Return} best$\_$CNN$\_$architecture\;
\end{algorithm}
The main goal of fitness evaluation is to give a quantitative measurement that determines which individuals qualify for serving as parent solutions. Algorithm 2 manifests the fitness evaluation algorithm in detail.

Firstly, a set of variables is initialized to zero (line 1). Following this, the process of fitness evaluation begins with randomly selecting an individual from the population and incrementing individual age, where individual age is the number of times the individual is selected for generating a CNN architecture (lines 2-4). Then, the individual is mapped to a CNN architecture, $C$, with all the hyperparameters being assigned during initialization (line 5). Next, from the process of genotype to phenotype, after being mapped every node in the Individual representation to small CNN, all small CNN assembled and made a big CNN, indicated as $C$. Then, weights and biases of the individual's last layer are set to the last layer of $C$ and the other parameters of $C$ are initialized by default based on the Keras framework (line 6). Then, for preventing thorough training of the architecture and a large number of epochs (\text{$>$}120 epochs inevitable for fully trained CNN), triggering high expenditure of computing resources, it is often determined 8 or 10 epochs for big image data sets. In our proposed method we set $k$, the number of epochs, to  ${k=1}$.

The CNN, ${C}$, is trained with the Adam optimizer (an extended version of SGD) on the training data, \text{D}$_{train}$, and the classification accuracy is calculated with the fitness evaluation data, \text{D}$_{val}$, (lines 7-9).
Subsequently, the fine-tuning of weights and biases associated with the final layer of the CNN architecture is accomplished through one epoch training process. Furthermore, in adherence to the Lamarckian concept, it is posited that knowledge acquired by each individual during their lifetime can be passed down to subsequent generations. As a consequence, the weights stored within the individual, which have undergone tuning, are considered as tuned weights. These tuned weights and biases, along with possible modifications introduced through operations such as weight-mutation and weight-crossover (depicted in algorithm 3), are then transmitted to the succeeding generation. Consequently, they are re-used to optimize initialization of the last layer weights and biases from the process of phenotype to genotype of the offspring at line 6 (line 10). Then, the evaluated fitness is compared to find the best fitness between the selected individuals, and the total fitness of selected individuals is computed (lines 11-15). After that, all modules that are genes of the individual, are updated based on the fitness of the individual (lines 16-18). Then, the counter is incremented (line 20). After that, the average fitness of the selected individuals is computed (line 21). Then, all modules and individuals which have not been used during the process of creating CNN architectures, are given a new chance to survive and evolve to the next generation by assigning the averaged fitness of all created CNN architectures (lines 22-31). Finally, the best CNN architecture is returned.

\subsection{Offspring Generation}
Offspring generation is the main part of the GA algorithm. Since all operators redefine and tune to pursue the goal of considering the domain of a given problem to find an optimized solution.

The process of offspring generating for the module and the individual is identical, with the exception of weighting-operators which are solely performed on the individual population because the module is determined as the gene of the individual and is not a real CNN. The process of offspring generation for individuals is illustrated in Algorithm 3, which is composed of two processes. The former is the process of structural-operators (lines 4-15) and the latter is the process of weighting-operators (lines 16-27). After finishing two processes for all individuals in the population, the new offspring-population, \text{Q}$_{t}$, is generated and returned (line 23). Firstly, \text{Q}$_{t}$ is initialized to zero. Then, during the process of offspring generation, for each species of population, the best ${k\%}$ of the species, $S$, are selected and added to the new offspring population (line 4). Then, two different individuals are selected as parents based on the tournament algorithm (lines 5-9). Next, the crossover operator is performed (lines 10-12), based on the principle of the NEAT method~\cite{14}. After that, structural mutation is performed (lines 13-15), consisting of some operations for the module population which is defined below: 
 \begin{itemize}

\item Adding a node, which is a layer when mapped to a real CNN architecture, between two uniformly selected nodes in the graph of the module.

\item The hyperparameters of a uniform node between the defined range are mutated as shown in table 1.
 \end{itemize}
For the individual, the mechanism of structural mutation consists of a number of actions defined below:
\begin{itemize}
\item Adding a node between two randomly selected nodes through the feedforward side which requires adding a concatenation layer when mapping the individual to real CNN.

\item Re-assigning pointer-module-species of a node to randomly selected module species.
 \end{itemize}
By performing structural operations, a new architecture is generated in which all parameters are initialized randomly. The rest of the algorithm is the process of weighting operators, and working on the weights and biases of the last layer of the CNN architecture, which is the most significant and effective layer for learning.
In order to do that, the fitness of the selected parent is computed, based on the accuracy measurement (lines 16-17). Then, a number, ${k}$, is uniformly generated in the range (${2/3;1}$) (line 18). Later, the computed fitness values of the parents are compared to find the best parent (line 19). 

As the goal of learning a neural network is to tune the neural network's weights, an individual with a better fitness value definitely has more tuned weights.  For that reason, in the process that we dub weighted crossover, the fittest parent should play a higher role in transmitting the genetic traits to the offspring. Therefore, the technique of weighted crossover is employed which allows ${k\%}$ of its weights of the last layer of the better parent and ${1-k\%}$ evolved weights of the last layer of the other parent to participate (lines 20-21 and 23-24). Note that, due to selecting parents from the same species, the similarity of every individual in terms of architecture is high, thus the tuned weights of the last layers of the architectures would be logically used for the crossover operator and employed to initialize weights of the last layer of the offspring. During the weights-mutation operation, a number is uniformly generated in the range $(0;1)$ (line 25), and the weights-mutation operation is performed on the offspring's weights if the generated number is lower than  ${t}$ (lines 26-27). For the weight-mutation operation, a random ${15\%}$ of the weights of the last layer of the offspring are perturbed with the value of 0.01 ${*}$ a generated selected number in the range $(0;1)$. Then the offspring are put into \text{Q}$_{t}$ (line 28) and finally, \text{Q}$_{t}$ is returned as the new population.

\begin{algorithm}
	\KwInput{the population individual or module $P_{t}$, all species of the population 
	$S_{pt}$.}
	\KwOutput{new offspring population $Q_{t}$.}
	$Q_{t} \gets 0$\;
	\While{$|Q_{t}| < |P_{t}|$ }
	{
		\ForEach{species as ${S}$ in \text{S}$_{pt}$ }
		{
			$Q_{t}$ += the 20\% best of ${S}$\;
			parent1 $\gets$ Randomly select two individuals from $S$ and then select 
			the one with the better fitness\;
			parent2 $\gets$ Repeat Line 5\;
			\While{parent1 == parent2 }
			{
				Repeat line 6\;
			}
		\textbf{end}
		$c$ $\gets$ generate uniformly number between (0,1)\;
		\If{$c<0.75 $}
			{
				child $\gets$ do \textbf{structural crossover} on parent1 and parent2\;
				child $\gets$ do \textbf{structural mutation} of the child\;
			}
		\Else
		{
			child $\gets$ do \textbf{structural mutation} uniformly on one of the parents\;
			fit$\_$P1 $\gets$ fitness evaluation of parent1\;
			fit$\_$p2 $\gets$ repeat line 15 for parent2\;
			$k$ $\gets$ uniformly generate a number between (2/3,1);
		}
			\textbf{end}
		}
	\textbf{end}
	\If{fit$\_$p1 $>$ fit$\_$p2}
	{
            child.W$_{lastLayer}$ $\gets$ $k/100 *$parent1.W$_{lastLayer}$ + $(1-k)/100 *$ parent2.W$_{lastLayer}$\;
            child.B$_{lastLayer}$ $\gets$ $k/100 *$parent1.B$_{lastLayer}$ + $(1-k)/100 *$ parent2.B$_{lastLayer}$\;
	}
	\Else
	{
            child.W$_{lastLayer}$ $\gets$ $k/100 *$parent2.W$_{lastLayer}$ + $(1-k)/100 *$ parent1.W$_{lastLayer}$\;
            child.B$_{lastLayer}$ $\gets$ $k/100 *$parent2.B$_{lastLayer}$ + $(1-k)/100 *$ parent1.B$_{lastLayer}$\;
	}
 
         $t$ $\gets$ generate a number uniformly selected in the range (0,1);
 
	\textbf{end}
		\If{$t$ $<$ 0.5 }
	{
		Uniformly select 15\% of the child weights and perturbing them with value of 
		0.01*uniformly number between (0,1)\;
		$Q_{t}$ $\gets$ child $\cup$ $Q_{t}$\;
	}
	\textbf{end}
	}
	\textbf{Return }$Q_{t}$
	\caption{Offspring Generation }\label{alg:three}
\end{algorithm}

\begin{table}
	\caption{Parameter configuration required by the LCoDeepNEAT algorithm.}
	\label{tab:one}
	\begin{tabular}{cc}
		\toprule
		Parameter & Value \\
		\midrule
		\multicolumn{2}{c}{Hyper-parameters}\\
		\midrule
		Number of outputs from a Convolution layer& [32, 80]\\
		Number of neurons in a fully connected layer& [128,800] \\
		Size of a Convolution kernel& [2,7]\\
		Number of layers  & [4,20]\\
		Range of drop out layer & [0.1,0.9]\\
		\bottomrule
	\end{tabular}
\end{table}

\section{Experiment Design}
In this section, to illustrate the empirical performance of the proposed method in discovering competitive neural network architectures, six widely used image classification datasets are described in subsection 4.1. Moreover, within subsection 4.2, the peer competitors are selected to compare with the proposed method. Furthermore, in subsection 4.3, we provide a comprehensive exposition of the parameter configurations utilized for conducting the experiments.

\subsection{Benchmark Datasets}
The benchmarks are categorized based on the nature of the classification objects they encompass. Specifically, three distinct categories are identified. The first category pertains to datasets comprised of handwritten images and their variations. Within this category, datasets such as MNIST, MNIST-RB (featuring random noise as background), and MNIST-BI (containing background images)~\cite{23} are included. The rationale behind selecting variants of the MNIST dataset lies in their heightened level of complexity, attributed to the presence of transformations such as rotations, the introduction of backgrounds, and other distracting attributes. Moving on, the second type of dataset focuses on fashion-related items and is referred to as the MNIST-Fashion dataset~\cite{24}. This dataset encompasses ten different types of fashion items, including clothing, footwear, and headwear. Lastly, the third type of dataset is employed for the identification of object shapes and comprises the Rectangles dataset and its counterpart, Rectangles-I, which incorporates images alongside the geometric shapes. Table 2 provides a comprehensive overview of the experimental datasets, encompassing information such as input size, number of classes, as well as the sizes of the training and test split samples.

\begin{table}
	\caption{evaluate the Overview of the datasets used to evaluate our experiments.}
	\label{tab:two}
	\begin{tabular}{cccc}
		\toprule
		Dataset&Input size&Classes&Training/Test\\
		\midrule
		MNIST & 28*28*1& 10& 50,000/10,000\\
		MNIST-RB & 28*28*1& 10& 12,000/50,000\\
		MNIST-BI & 28*28*1& 10& 12,000/50,000\\
		Rectangles & 28*28*1& 2& 1,200/50,000\\
		Rectangles-I & 28*28*1& 2& 1,200/50,000\\
		MNIST-Fashion & 28*28*1& 10& 60,000/10,000\\
		\bottomrule
	\end{tabular}
\end{table}

\subsection{Peer Competitors}
In order to illustrate the effectiveness and efficiency of the proposed method, state-of-the-art methods are chosen as the peer competitors. Notably, the peer competitors are selected from two different categories. The former category is state-of-the-art CNNs which are manually designed, including LeNet~\cite{25}, NNET~\cite{26}, and SAA-3 ~\cite{26}. In addition, 10 peer competitors on the second category of the dataset are collected from the FASHION dataset homepage~\cite{27}. They are 2C1P2F+Dropout, 2C1P, 3C2F, 3C1P2F+Dropout, GRU+SVM+Dropout, GoogleNet~\cite{20}, AlexNet~\cite{2}, SqueezeNet-200~\cite{28}, MLP 256-128-64, and VGG16~\cite{3}. The latter are state-of-the-art CNNs which are automatically generated, consisting of IPPSO~\cite{21}, MBO-ABCFE~\cite{29}, GeNet~\cite{30}, DNNCOCA~\cite{31}, psoCNN~\cite{22}, sosCNN~\cite{33}, CoDeepNEAT~\cite{34}, and SEECNN~\cite{32}.

\subsection{Parameter Setting}
In the training phase of each individual, the Adam optimization algorithm, a variant of SGD, is employed with a learning rate of 0.001. This choice of optimization algorithm and learning rate allows for the acceleration of the evaluation process and the capture of the performance trends exhibited by a CNN. To conform with commonly accepted practices in the machine learning community, a validation set is created by randomly selecting \text{20\%} of the samples from the training set. The batch size is set to 108, determining the number of samples processed in each iteration. For each benchmark dataset, the population size of individual and population modules is established at 25 and 30, respectively. A total of 10 generations are executed, and within each generation, 15 candidate solutions are considered. It is worth noting that larger population sizes and an increased number of generations would theoretically lead to improved performance, albeit at the expense of greater computational resources.
Regarding the parameters of the CNN layer, a stride of (1,1) is set for the convolutional layer, while the pooling layer adopts the same kernel size. The padding type for both the convolutional and pooling layers is consistently defined as "SAME," ensuring that the input and output dimensions remain unchanged. Once the proposed method concludes, the individual exhibiting the highest fitness value is selected, and a subsequent training phase is conducted, consisting of 100 epochs, using the original training set.
The implementation of LCoDeepNEAT leverages the Keras deep learning framework. All experiments are conducted on a computer equipped with a single GPU card, specifically the Nvidia GeForce GTX 1080 Ti model. To ensure the reliability and authenticity of the results, LCoDeepNEAT is independently executed 10 times for each benchmark dataset.
\section{Experimental Results and Analysis}
In this section, we commence by presenting a comprehensive overview of the comparative outcomes between the proposed methodology and the selected peer competitors. Then, we examine the performance of the proposed method without using lamarckian idea and evolving last layer in comparison with the originally proposed method.

\subsection{Overall Results}

\begin{table*}
 \centering
	\caption{The best and mean classification error rates of LCoDeepNEAT and its
 peer competitors on the MB, MBI, MRB, RECT and RI datasets. An entry of ``\textendash'' indicates
 that the information was not reported or is not known to us.
 Much of this table was based on that presented in ~\cite{32}.}
	\label{tab:three}
	\begin{tabular}{cccccc}
		\toprule
		Classifier &MB&MBI& MRB&RECT&RI \\
		\midrule
		\multicolumn{6}{c}{Hand-crafted architectures}\\
		\midrule
		LeNet-1~\cite{25}& 1.70\% &9.80\%&7.50\%&-&16.92\%\\
		NNet~\cite{26}& 4.96\% &27.41\%&20.04\%&7.16\%&33.20\%\\
		SAA-3~\cite{26}& 3.46\% &23.00\%&11.285\%&2.41\%&24.05\%\\
		\midrule
		\multicolumn{6}{c}{Evolutionary algorithms for architecture generation}\\
		\midrule
		IPPSO (best)~\cite{21}&1.13\% &-&-&-&-\\
  		IPPSO (mean)~\cite{21}&1.21\% &-&-&-&-\\
		MBO-ABCFE (best)~\cite{29}&0.34\% &-&-&-&-\\
		GeNET (best)~\cite{30}&0.34\% &-&-&-&-\\
		DNN-COCA (best)~\cite{31}&1.30\% &-&-&-&-\\
  		CoDeepNEAT (best)~\cite{34}&0.77\% &-&-&-&-\\
		SEECNN (best)~\cite{32}& 0.79\% &4.06\%&2.44\%&\textbf{0.00}\%&2.18\%\\
		psoCNN (best)~\cite{22} & \textbf{0.32\%} &1.90\%&1.79\%&0.03\%&2.22\%\\
            psoCNN (mean)~\cite{22} & 0.44\% &2.40\%&2.53\%&0.34\%&3.94\%\\
		sosCNN (best)~\cite{33} &0.38\% &1.68\%&\textbf{1.49\%}&\textbf{0.00\%}&1.57\%\\
  		sosCNN (mean)~\cite{33}&0.40\% &1.98\%&1.89\%&\textbf{0.01\%}&2.37\%\\
  		\textbf{LCoDeepNEAT (best)} &0.33\% &\textbf{1.02\%}&1.52\%&\textbf{0.00\%}&\textbf{1.40\%}\\
    	\textbf{LCoDeepNEAT (mean)} &\textbf{0.39\%} &\textbf{1.30\%}&\textbf{1.70\%}&\textbf{0.01\%}&\textbf{1.90\%}\\
		\bottomrule
	\end{tabular}
\end{table*}

In this subsection, we present the best and mean classification error rates of CNNs evolved by LCoDeepNEAT on the six datasets. These results are provided in Table 3 and Table 4 for reference. The results show that the LCoDeepNEAT method outperforms all hand-crafted methods and obtains the best mean error rates in comparison with competitive NAS methods.

In the context of the MNIST benchmark, as delineated in Table 3, the psoCnn technique demonstrated the top-1 error rate of 0.32\%. In contrast, our proposed method, LCoDeepNEAT, achieved an error rate of 0.33\%, whereby the disparities between the two results are negligible. Turning our attention to the MNIST-BI dataset, the sosCNN
method exhibited an error rate of 1.68\%, whereas our method, LCoDeepNEAT, not only attained the lowest best error rate of 1.02\% but also achieved the lowest mean error rate of 1.30\%. This represents a notable enhancement of 0.66\% and 0.38\%, respectively, when compare to sosCNN. Regarding the MNIST-RB dataset, LCoDeepNEAT recorded the second-best error rate of 1.52\%.

In the case of the Rectangles dataset, our approach, sosCNN, and SEECNN all
demonstrated the minimal error rate of 0.00\%, while psoCNN recorded the highest error rate of 0.03\% among the NAS methods. Lastly, for the Rectangles-I dataset,
sosCNN emerged with the second-best error rate of 1.57\%, whereas our method achieved the best error rate  of 1.40\%.

Moreover, the findings presented in Table 3 unequivocally establish the superior
performance of LCoDeepNEAT over all other NAS techniques, as evidenced by its consistently lower mean error rates across diverse datasets. This outcome highlights the advantages derived from incorporating evolving partially weights of the last layer, lamarckian idea and constraining the search space for architectural exploration. The empirical evidence strongly supports the assertion that the proposed search strategy accelerates the process of uncovering optimal solutions.

In this study, we also carried out experiment on the MNIST-Fashion dataset. The experimental results, demonstrating the best classification performance of various methods, are presented in Table 4. Remarkably, the LCoDeepNEAT method surpasses all
handcrafted methods in terms of error rates, even outperforming renowned architectures such as GoogleNet, AlexNet, and VGG-16. Additionally, among the NAS methods, the SEECNN method achieved a minimum classification error of 5.38\% with the highest parameter count, 15.9 million parameters, while LCoDeepNEAT attained a 6.21\% error rate with the lowest parameter count of 1.2 million parameters. This experiment illustrates that finding a trad-off between classification accuracy and architectural complexity is extremely intricated for NAS methods. However, LCoDeepNEAT demonstrates its ability to strike an optimal trade-off, discovering optimized architectures with comparable accuracy.

\begin{table}
	\caption{LCoDeepNEAT classification errors (\%) for the MNIST-Fashion benchmark 
	dataset compared to different models. An entry of ``\textendash'' indicates that the information 
	was not reported or is not known to us. Much of this table was based on that presented in ~\cite{32}.}
 \centering
	\label{tab:four}
	\scalebox{0.8}{
	\begin{tabular}{cccc}
		\toprule
		Classifier &Error (\%)&parameters (million)& epochs\\
		\midrule
		\multicolumn{4}{c}{Hand-crafted architectures}\\
		\midrule
		2C1P2F+Drouout~\cite{27}& 8.40(+)&3.27&300\\
		2C1P~\cite{27}&7.50(+) &100K&30\\
		3C2F~\cite{27}& 9.30(+) &-&-\\
		3C1P2F+Dropout~\cite{27}& 7.40(+) &7.14&150\\
		GRU+SVM+Dropout~\cite{27}& 10.30(+) &-&100\\
		GoogleNet~\cite{20}& 6.30(+) &101&-\\
		AlexNet~\cite{2}& 10.10(+)&60&-\\
		SqueezeNet-200~\cite{28}& 10.00(+) &\textbf{500K}&200\\
		MLP 256-128-64~\cite{27}& 10.00(+)&41K&25\\
		VGG16 \cite{3}& 6.50(+)&26&100\\
		\midrule
		\multicolumn{4}{c}{Evolutionary algorithms for architecture generation}\\
		\midrule
		EvoCNN~\cite{13}&5.47&6.68&100\\
		SEECNN~\cite{32}&\textbf{5.38}&15.9&100\\
		sosCNN~\cite{33}&5.68&2.30&100\\
		\textbf{LCoDeepNEAT}&6.21 &\textbf{1.2}&100\\
		\bottomrule
	\end{tabular}
}
\end{table}

We present the efficiency of utilizing lamarckian idea and evolving the last layer of candidate solutions. These combined strategies in LCoDeepNEAT yield a noteworthy increase in classification accuracy for all datasets during evolutionary process, ranging from 2\% to 5.6\%. For example, figure 3 illustrates classification accuracy of our proposed method with and without using these strategies on MNIST-BI dataset in 10
generations on one epoch.

\subsection{Performance with Weight Evolving}
In this section, we examined LCoDeepNEAT with just evolving partially last layer weights without using lamarckian idea in one epoch on all datasets. The results are classification accuracies in every generation during evolution. Our experiments show that this strategy increases classification accuracy in every generation by 0.4\% to 0.8\%.

\subsection{Performance with Constraint on Architecture Search Space}
LCoDeepNEAT demonstrates the ability to identify optimal  candidate solutions in each generation, facilitated by a meticulously constrained space for architectures
within the candidate pool. To substantiate our assertion, we conducted experiments where the last layer was not evolved, and lamarckian idea  was not utilized. The outcomes revealed that our method succeeded in identifying the optimal architectures for all datasets. For Instance, on the MNIST-BI dataset, LCoDeepNEAT with an only constraint on architecture search space achieves an impressive classification error rate of 1.63\%, and the originally proposed method obtained a lower error rate at 1.02\%, whereby the disparities between the two results are 0.61\%.

\begin{figure}[ht]
	\centering
	\includegraphics[width=\linewidth]{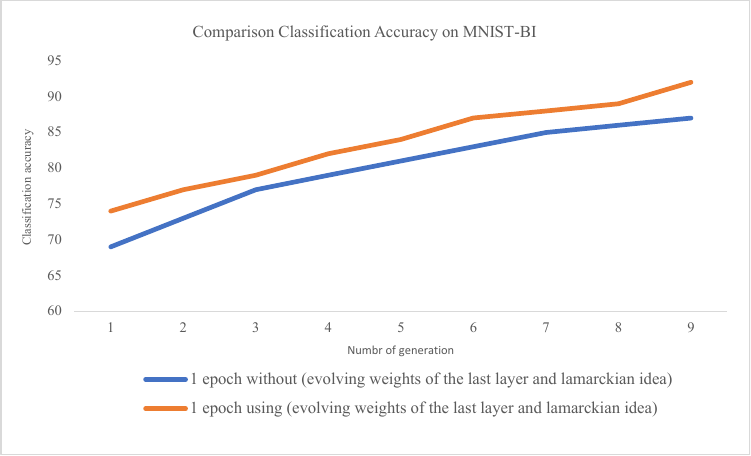}
	\caption{Comparison classification accuracy of LCoDeepNEAT with and without using evolving
weights of the last layer and lamarckian idea on MNIST-BI in 10 generations.}
        \label{fig-3}
\end{figure}

\section{Conclusion and Future Work}
In this paper, we introduce LCoDeepNEAT, an efficient GA method, that facilitates
the simultaneous evolution of CNN architectures and their last layer weights.
LCoDeepNEAT method utilized a direct encoding strategy that shows the architecture
as a graph and stores the weights and bias of the last layer connections in genotype. We
also employ the Lamarckian idea to transfer the tuned weights (gradient information)
to the weight component of the network's genotype during evolution.

LCoDeepNEAT demonstrates a notable improvement in the classification accuracy of candidate solutions throughout the evolutionary process, ranging from 2\% to 5.6\%. This outcome underscores the efficacy and effectiveness of integrating gradient information and evolving the last layer of candidate solutions within our method. 

In addition, LCoDeepNEAT, with just evolving partially last layer weights without using gradient information in one epoch on all datasets, increases classification accuracy of candidate solutions during evolution by 0.4\% to 0.8\%. These results highlight the potential of evolving the last layer weights as a standalone strategy within the LCoDeepNEAT method.

Moreover, LCoDeepNEAT expedites the process of evolving by imposing restrictions on the architecture search space, specifically targeting architectures comprising just two fully connected layers for classification.

LCoDeepNEAT is tested and compared with eight recent NAS methods and more
than twelve hand-crafted methods on six widely used datasets. LCoDeepNEAT in terms
of best error rate demonstrated superior classification performance compared to all
hand-crafted methods across all datasets. It also surpasses some NAS methods as well.
Furthermore, in terms of mean error rates, LCoDeepNEAT outperforms all NAS methods across the evaluated datasets.
Given the inherent characteristics of LCoDeepNEAT, the utilization of trained modules within the evolutionary process becomes feasible. Consequently, in our future endeavors, we plan to consider the evolution of module weights, incorporating gradient information, assessing their influence on the discovery of optimized architectures, and fine-tuning the weights of architectures.

\bibliography{IEEEtran/bare_conf}
\bibliographystyle{IEEEtran}

\end{document}